\documentclass[krantz1,ChapterTOCs]{krantz} 
\usepackage{amssymb}
\usepackage{amsmath}
\usepackage{graphicx}
\usepackage{subfigure}
\usepackage{makeidx}
\usepackage{multicol}
\usepackage{multirow}
\usepackage{makecell}

\usepackage{cite}
\usepackage{times}
\usepackage{graphicx}
\usepackage{gensymb}
\usepackage{booktabs}
\usepackage[style=base]{caption}
\usepackage{mathtools}
\usepackage{pifont}
\usepackage{xspace}
\usepackage{xfrac}
\usepackage{microtype}
\usepackage[table]{xcolor}

\usepackage{float}  
\usepackage{wrapfig} 
\usepackage{acronym} 
\usepackage{booktabs} 
\usepackage{adjustbox} 
\usepackage{comment}
\usepackage{tabulary}
\usepackage{siunitx}

\makeindex

{\par\addvspace{6pt}\normalfont {\bfseries #1}\hskip\labelsep\ignorespaces\itshape}
{\par\addvspace{6pt}}

\begin{document}
\frontmatter

\title{Krantz Template} 
\author{Yours Truly}


\tableofcontents
\let\cleardoublepage\clearpage
\mainmatter

\chapterauthor{David Unger, 
Nikhil Gosala,
Varun Ravi Kumar, 
Shubhankar Borse, 
Abhinav Valada,
Senthil Yogamani}{\emph{Keywords}: Semantic Segmentation, 3D Object Detection, Fisheye Cameras, Autonomous Driving, Bird's Eye View.} 
\chapter{Multi-camera Bird's Eye View Perception for Autonomous Driving}
\newcommand{\secref}[1]{Section~\ref{#1}}
\renewcommand{\eqref}[1]{Equation~(\ref{#1})}
\newcommand{\figref}[1]{Figure~\ref{#1}}
\newcommand{\tabref}[1]{Table~\ref{#1}}
\newcommand{\red}[1]{\textcolor{red}{#1}}
\acrodef{IPM}{Inverse Perspective Mapping}
\acrodef{FoV}{Field-of-View}
\acrodef{EV}{Ego View}
\acrodef{PV}{Perspective View}
\acrodef{BEV}{Bird's Eye View}
\acrodef{MLP}{Multilayer Perceptron}
\acrodef{IoU}{Intersection over Union}
\acrodef{mIoU}{Mean Intersection over Union}
\acrodef{TP}{True Positive}
\acrodef{FP}{False Positive}
\acrodef{FN}{False Negative}
\acrodef{AP}{Average Precision}
\acrodef{mAP}{Mean Average Precision}
\acrodef{NDS}{nuScenes Detection Score}
\acrodef{AUC}{Area Under the Curve}
\acrodef{MP}{Megapixel}
\acrodef{ADAS}{Advanced driver-assistance systems}
\acrodef{OEM}{Original Equipment Manufacturer}
\acrodef{CNN}{Convolutional Neural Network}
\acrodef{CV}{Computer Vision}

\acrodef{LSS}{Lift-Splat-Shoot}
\acrodef{CVT}{Cross-View-Tranformer}
\acrodef{BiFPN}{Bi-directional Feature Pyramid Network}
Most automated driving systems comprise a diverse sensor set, including several cameras, Radars, and LiDARs, ensuring a complete $360^{\circ}$ coverage in near and far regions.  
Unlike Radar and LiDAR, which measure directly in 3D, cameras capture a 2D perspective projection with inherent depth ambiguity. However, it is essential to produce perception outputs in 3D to enable the spatial reasoning of other agents and structures for optimal path planning. The 3D space is typically simplified to the \ac{BEV} space by omitting the less relevant Z-coordinate, which corresponds to the height dimension.\par

The most basic approach to achieving the desired \ac{BEV} representation from a camera image is \ac{IPM}, assuming a flat ground surface. Surround vision systems that are pretty common in new vehicles use the \ac{IPM} principle to generate a \ac{BEV} image and to show it on display to the driver. However, this approach is not suited for autonomous driving since there are severe distortions introduced by this too-simplistic transformation method.\par

More recent approaches use deep neural networks to output directly in \ac{BEV} space. These methods transform camera images into \ac{BEV} space using geometric constraints implicitly or explicitly in the network. As \acp{CNN} has more context information and a learnable transformation can be more flexible and adapt to image content, the deep learning-based methods set the new benchmark for \ac{BEV} transformation and achieve state-of-the-art performance.\par

First, this chapter discusses the contemporary trends of multi-camera-based DNN (deep neural network) models outputting object representations directly in the \ac{BEV} space. Then, we discuss how this approach can extend to effective sensor fusion and coupling downstream tasks like situation analysis and prediction. Finally, we show challenges and open problems in \ac{BEV} perception.\par
\section{Introduction}
\label{intro}

Scene perception and understanding form the basis for automated driving, and the quality of scene understanding dictates the performance of downstream tasks such as path planning and control. Accurate scene understanding is often challenging in urban environments due to complex interactions with actors and challenging driving scenarios, including construction areas or intersections~\cite{uricar2019challenges, dhananjaya2021weather, klingner2022detecting}. Camera-based perception has been a critical component of autonomous driving perception systems including a wide variety of perception tasks like depth estimation~\cite{kumar2020unrectdepthnet, kumar2021svdistnet, kumar2018near}, motion estimation~\cite{rashed2019motion, rashed2021bev}, image restoration \cite{uricar2019desoiling}, semantic segmentation \cite{dutta2022vit} and multi-task learning \cite{leang2020dynamic, sistu2019real}. They are frequently used due to their high angular resolution, dense visual representation at high distances, and cheap manufacturing and integration costs. In a multi-sensor setting consisting of RADARs, LiDARs, and ultra-sonic sensors, a successful multi-sensor camera fusion becomes a key research and engineering problem to focus on~\cite{borse2023x, mohapatra2021limoseg}. In this article, we shall study the various representations used within Camera-based perception to better align with 3D-perception outputs obtained from any other sensors mentioned above. In the context of automated driving, two output domain representations are popular in 3D perception systems : 3D detection-based and 2D-occupancy grid-based. Objects are represented by oriented bounding boxes in 3D, while a Cartesian discrete grid is utilized in the latter. Based on which representation is utilized an adapted fusion with other sensors can be achieved.\par

Camera-based systems has depended on two types of image representations : \textit{Perspective View (PV) representation} and \textit{Bird's eye View (BEV) representation}. \textit{\ac{BEV} representation} have gained much attention due to their efficacy in different parts of the automated driving pipeline. The \ac{PV} representation depicts the scene from the viewpoint of a sensor mounted on an automated vehicle. It captures the height dimension of the scene, which makes it extremely useful when the vehicle drives through overhanging regions such as tree canopies, bridges, and tunnels. Further, \ac{PV} allows for the inference of road rules via traffic light and traffic sign detection, as well as the intention of other agents via blinker and brake-light detection for vehicles and human pose estimation for pedestrians. It also enables the detection of appearance-based social cues, such as a person's age and the presence of a visual disability, which can be used to develop socially-aware automated driving systems. The above characteristics make PV extremely suitable for scene-understanding tasks such as segmentation and object detection. The main goal is to analyze and describe the attributes of various elements in the scene. However, it is essential to have these perceived objects in 3D for downstream automated driving tasks.\par

\begin{figure}[!ht]
  \includegraphics[width=\linewidth]{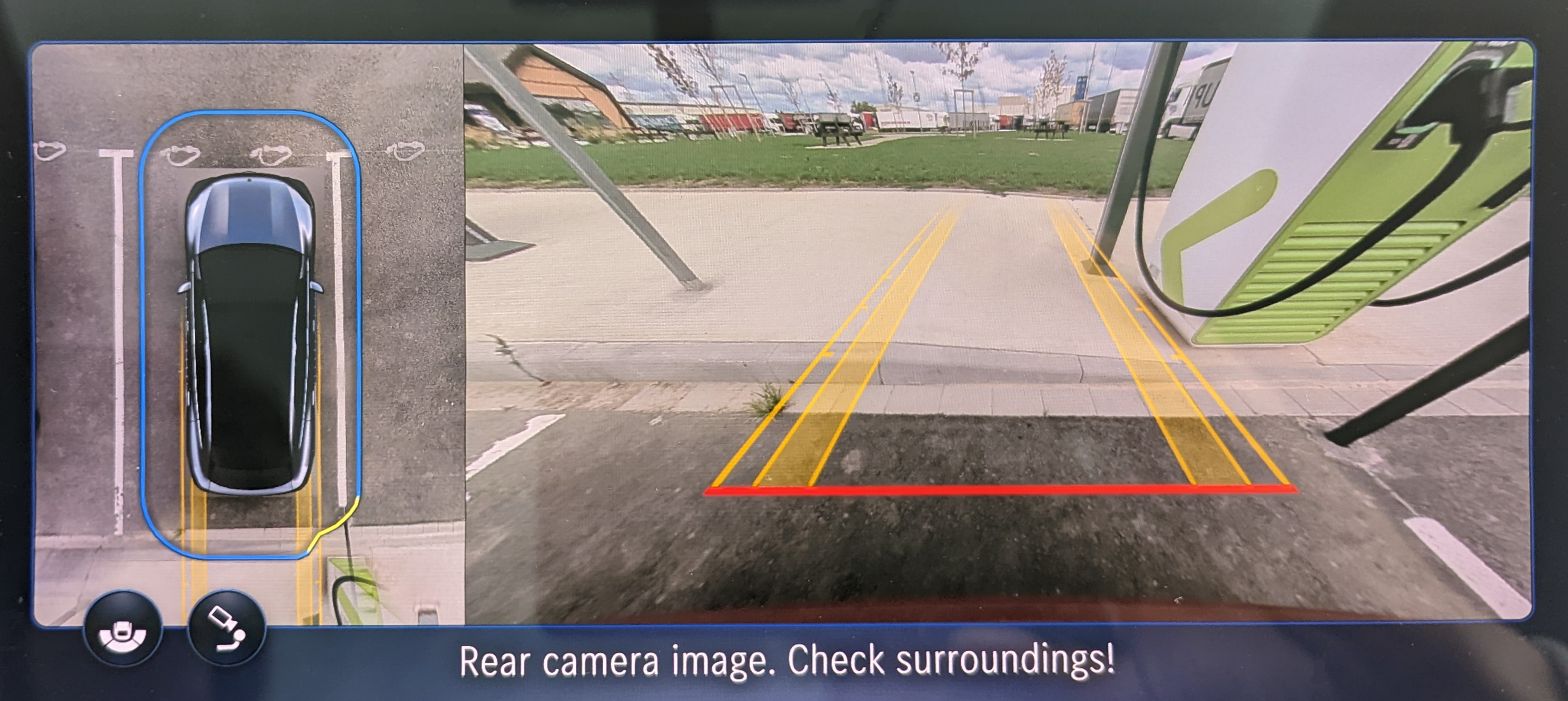}
  \caption{Vehicle manufacturers offer to surround vision systems to show a top-down representation of the environment, which is helpful for humans but inappropriate for autonomous driving since it does not provide an accurate scale. Further, it only shows color but does not provide information about the image context.}
  \label{fig:surroundview}
\end{figure}

On the other hand, the BEV representation depicts the scene from the viewpoint of a downward-facing virtual orthographic camera placed above the automated vehicle. Historically, surround vision systems commonly used the \ac{IPM} principle to generate a \ac{BEV} image and to show it on display to the driver, as depicted in \figref{fig:surroundview}. It captures the scene's depth proportional to the metric scale, which allows it to be directly used for distance-sensitive tasks such as collision avoidance. It also can explicitly capture occlusions in the scene, allowing subsequent tasks, such as path planning and control, to handle the ambiguity associated with such regions gracefully. Lastly, being an orthographic projection, BEV representation does not suffer from perspective distortion inherent to PV, simplifying the representation and processing of lane geometry and road markings. These characteristics make BEV representation apt for decision-based tasks such as trajectory estimation and control, where the main aim is to safely interact with the static and dynamic elements of the scene. Thus, both PV and BEV representations form an integral part of automated driving and are crucial for an automated vehicle's safe and efficient operation.\par

PV and BEV representations are typically employed at distinct stages of the perception pipeline, with PV mainly used in the earlier stages for tasks such as segmentation and tracking. In contrast, BEV is employed in subsequent sensor fusion and path planning tasks. Thus, there is a disconnect within the pipeline since the generation of BEV representation requires a depth estimate that cannot be directly obtained from the PV representation. Multiple approaches address this disconnect by either explicitly predicting the scene's depth using depth estimation networks~\cite{ming2021deep} or by extracting depth information from range-based sensors such as LiDARs~\cite{premebida2016high}. These depth estimates are then combined with the intermediate outputs from the PV to generate the required BEV representations. These multi-stage approaches, however, generate sub-optimal BEV representation due to scale inconsistencies in depth estimation networks and the sparsity of range-based sensors. Many recent works have proposed various deep learning-based approaches to generate BEV representations directly from PV images, following an end-to-end learning strategy to alleviate this limitation. These approaches learn the complex characteristics of PV-BEV mapping using neural networks, thus generating highly accurate representations in the BEV. For instance, VPN~\cite{pan2019cross} uses two MLP layers to learn the PV-BEV mapping, Cam2BEV~\cite{reiher2020sim2real} augments the output of IPM with a learnable transformer, LSS~\cite{philion2020lift} predicts a depth distribution to lift the PV features into the BEV space, PanopticBEV~\cite{gosala2022bird} uses dual transformers to independently transform vertical and flat regions in the PV image to BEV, and BEVFormer~\cite{li2022bevformer} employs self-attention-based transformers to generate the required PV-BEV transformation.\par
\begin{figure}[!ht]
  \includegraphics[width=\linewidth]{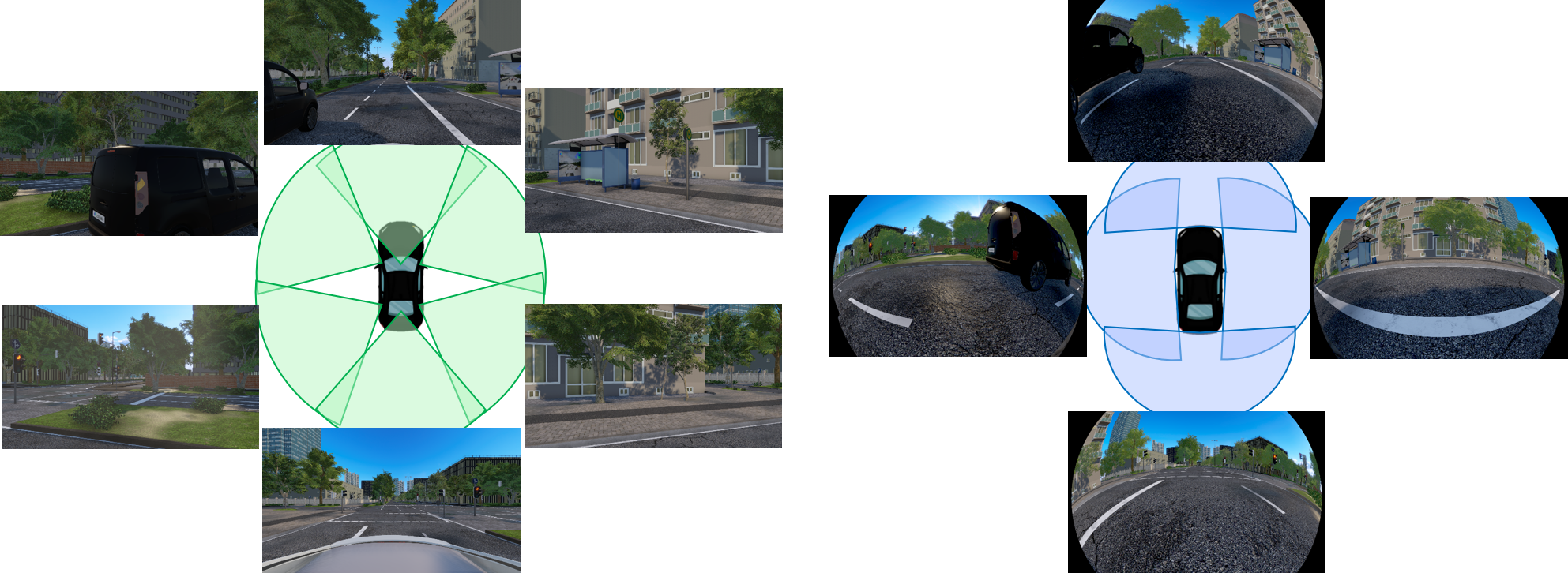}
  \caption{A typical camera sensor set of an automated vehicle comprises several pinhole cameras and typically four fisheye cameras.}
  \label{fig:od-fov}
\end{figure}
With the development of automotive platforms equipped with multiple cameras generating a 360\,° view around the vehicle, it has become imperative to address the concerns pertaining to the optimal fusion of data obtained from these surround-view cameras. In automated driving systems, the cameras are designed to have both far-field and near-field 360\,° view around the vehicle as shown in Figure \ref{fig:od-fov}. Accordingly two strategies, namely, \textit{late fusion} and \textit{early fusion}, have been proposed in the literature. In late fusion, data from each camera is first processed independently to generate the required output. The outputs are then manually fused into a coherent representation via a post-processing step to generate the surround-view output.
In contrast, approaches following the early fusion paradigm fuse data from multiple cameras within the model and then coherently process the fused data to generate the required surround-view output. Early fusion approaches typically have several advantages over their late fusion counterpart. To name a few: The detection of large objects that often span multiple cameras becomes easier as all necessary information is readily available in the fused feature space.
Re-identification of objects across cameras while performing object tracking becomes redundant.
Predictions in regions where one camera overlaps another become trivial as no heuristic-based post-processing step is needed to handle contradictory predictions.
In this direction, multiple approaches have been proposed that leverage information from surround view cameras to generate coherent predictions in the BEV. However, early fusion comes with an increased computational and memory cost in terms of the extremely large intermediate feature space, which often limits the resolution of the input and output spaces. With more cameras becoming an integral part of an automated vehicle's sensor suite, unifying all cameras into a coherent multi-camera pipeline is essential.\par
{In this chapter, we present the various deep-learning-based techniques developed for the task of single- and multi-view BEV perception. We first introduce the various tasks pertaining to BEV perception in \secref{sec:tasks} and then elaborate on the architectural details of various algorithms developed to address these tasks in \secref{sec:arch}. Furthermore, we present an overview of the datasets and metrics used for BEV perception in \secref{sec:datasets}. Finally, we discuss the various challenges and open problems pertaining to various stages of the BEV perception pipeline in \secref{sec:challenges}.}
\section{Perception Tasks} 
\label{sec:tasks}

Perception and planning methods have developed independently in different communities. They have led to an inherent disconnect between PV and BEV, hindering the quest for developing a coherent end-to-end automated driving pipeline. This has led to work producing BEV representations directly from PV images in the past few years. This development has been fueled by the success of deep learning-based approaches, which enabled more end-to-end learned complex mappings from large amounts of data. Accordingly, several authors have proposed learning-based solutions to transform PV images into BEV. In most approaches, the PV-BEV transformation occurs within the feature space, creating a comprehensive perception framework where multiple tasks, such as segmentation, tracking, and prediction, can be performed in the BEV space. In this section, we present two popular BEV perception tasks, namely, 3D Object Detection and BEV Segmentation.\par
\subsection{3D Object Detection in Cameras}

\begin{figure}[!ht]
  \centering  
  \includegraphics[width=10cm]{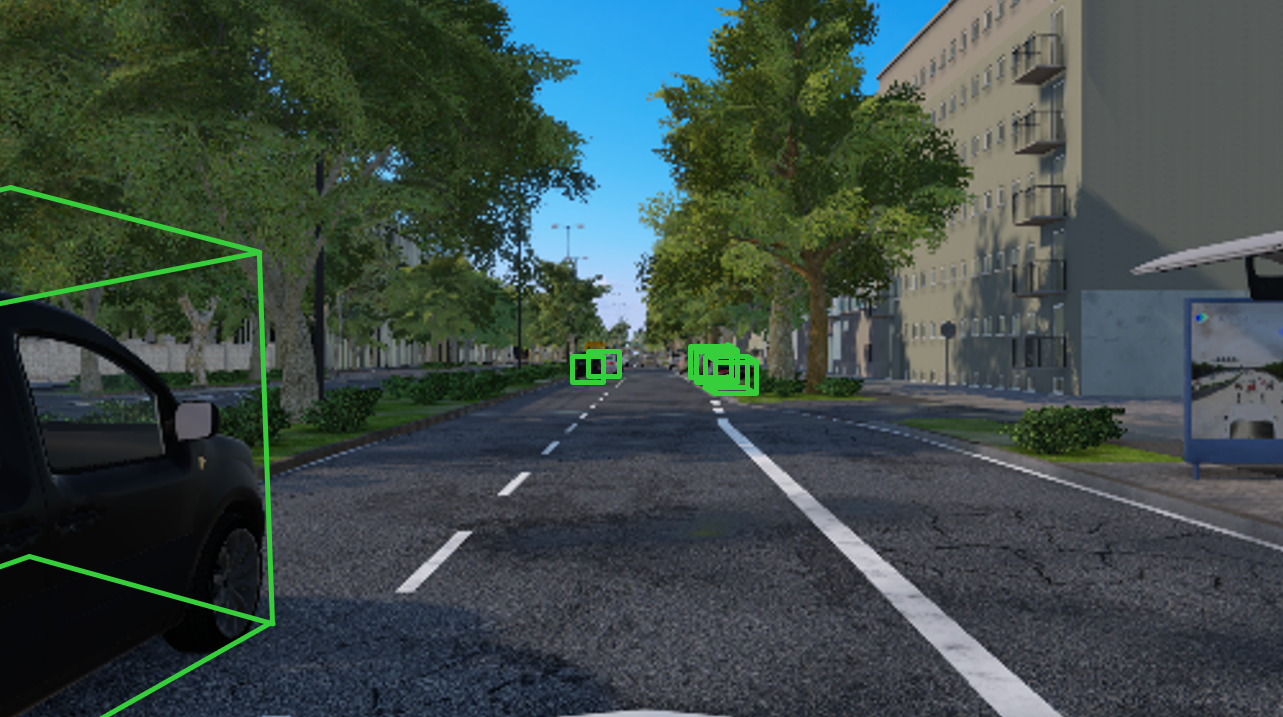}
  \caption{3D object detection has the target to output an object's location, size, rotation, and class.}
  \label{fig:3dod}
\end{figure}

3D Object Detection refers to the task of identifying objects in the scene and estimating 3D bounding box parameters around them. This task encompasses two significant schools of thought: image plane-based object detection and BEV space-based object detection, which differ in the input domain over which the detection head performs object detection. \figref{fig:3dod} illustrates the 3D object detection output by projecting the bounding box predictions onto the corresponding PV image. Image plane-based object detection pipelines comprise approaches that directly regress 3D information of the object, such as 3D size, 3D position, and orientation, onto the input PV image~\cite{wang2021fcos3d}. These approaches are typically affected by the perspective distortion of PV images which causes similar objects at different distances to appear at different scales, which prevents the network from learning meaningful object representations~\cite{roddick2018orthographic}. The lack of object consistency also results in poor 3D localization performance, which forces such approaches to rely on explicit depth supervision to improve their performance. Lastly, the detection task is required to be robust to object occlusions.\par

In contrast, BEV space-based camera-based object detection consists of approaches that perform object detection by estimating 3D bounding box coordinates by evaluating transformed BEV features from the camera front view. The BEV space represents the world using the orthographic projection, which addresses issues on scale inconsistency and object occlusion of perspective projection. Further, the transformation from PV to BEV enables better spatial and geometric reasoning resulting in better object localization and, accordingly, a better overall performance~\cite{roddick2018orthographic, huang2022bevdet}.\par
\subsection{BEV Segmentation with Cameras}

BEV segmentation in cameras is the task of evaluating a metric semantic occupancy grid map using a single monocular image or a cocoon of images around the ego-vehicle. Semantic segmentation on point clouds or LiDARs is easier to define since point clouds provide a native BEV presentation. At the same time, an 2D/3D occupancy representation depends strongly on the projective geometry and camera parameters. The BEV-based representations are useful upstream for planned and scene interpretation models. These frameworks frequently simplify the autonomous vehicle's movement into the XY plane with a single angle representing orientation within the plane. Whereas 3D detection provides oriented bounding boxes with more degrees of liberty.\par

\begin{wrapfigure}{R}{0.5\textwidth}
  \begin{center}
    \includegraphics[width=0.45\textwidth]{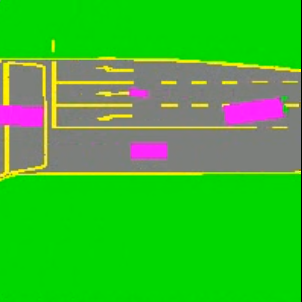}
    \caption{\ac{BEV} segmentation provides the occupancy, semantic label, and precise information about which thing or stuff is located in space.}
    \label{fig:bev_segmentation}
  \end{center}
\end{wrapfigure}

BEV segmentation-based perception systems generally involve (i) BEV semantic segmentation, which assigns a semantic class label to each pixel in the output space, (ii) BEV instance segmentation which distinguishes between the different instances of an object, and (iii) BEV panoptic segmentation which combines BEV semantic and instance segmentation to generate a coherent output that semantically distinguishes regions in the BEV space while simultaneously discerning between instances of an object~\cite{gosala2022bird}. The BEV segmentation output is typically represented using a rasterized grid representing the vehicle's surroundings. The grid cells are often uniformly spaced, and each grid cell contains the segmentation label about the corresponding real-world location. \figref{fig:bev_segmentation} graphically represents a typical BEV panoptic segmentation output where the road class is depicted in gray, road markings in yellow, grass and shrubs using green, and distinct instances of the vehicle class are colored differently.\par
BEV segmentation forms an integral part of the scene understanding pipeline and is critical for the proper functioning of various subsequent tasks. For example, BEV panoptic segmentation facilitates the segmentation of static classes, such as roads and sidewalks. BEV panoptic segmentation also Identifies object instances belonging to dynamic classes is critical for downstream tasks such as path planning and trajectory estimation. The BEV instance labels also assist in tracking vehicles and pedestrians over time, which can help the model reason about occlusions and can positively influence the decision-making process of the autonomous vehicle. Further, BEV segmentation allows for the creation of instantaneous HD maps, which can be accumulated over time to generate temporally coherent HD maps. These temporally coherent maps can, in turn, be used in creating HD maps for new locations or can assist in the maintenance of previously annotated HD maps.\par

The BEV segmentation output is characterized by two important design parameters: grid size and grid resolution of the rasterized BEV grid. The grid size defines the number of grid cells and determines the spatial size of the \ac{BEV} segmentation output. The larger the grid size, the higher the information that can be represented in the segmentation output. However, a larger grid size increases the computational complexity, which subsequently increases the computational requirements as well as the runtime of the model. In contrast, grid resolution determines the real-world size of a single grid cell. Given a fixed grid size, a finer resolution allows the BEV grid to capture finer details, such as elements belonging to lane markings and pedestrians, at the expense of a smaller real-world area. In comparison, a coarser grid resolution increases the real-world area covered by the BEV grid at the cost of a decreased segmentation granularity. Some works in the literature use a grid size of $200 \times 200$ cells with each having a resolution of \SI{0.5}{\meter}, thus capturing a total real-world area of $\SI{100}{\meter} \times \SI{100}{\meter}$~\cite{philion2020lift, hu2021fiery, xie2022m2bev}. However, a human covers an area of less than one grid cell at such coarse resolutions, which makes it unsuitable for city driving. Other works use a finer grid resolution of \SI{0.075}{\meter} with a $\sim 4\times$ larger grid size ($768 \times 704$) to segment small objects while capturing a range of ~\SI{50}{\meter} in each direction~\cite{gosala2022bird}. Due to the large segmentation output,~\cite{gosala2022bird} has a higher runtime than their coarse-resolution counterparts, making it unsuitable for time-sensitive applications. In general, a trade-off must be made between the spatial size and grid resolution of the segmentation output, as well as the runtime of the overall model. Given the large impact of each of these parameters not only on each other, but also on the applicability of the overall segmentation model in the real-world, it is crucial to incorporate them into the set of fundamental design decisions for BEV perception. In the ideal case, the segmentation output would have a large grid size, a fine grid resolution while the segmentation model would have a very low runtime. Since most autonomous driving applications are bounded by strict runtime constraints for safety purposes, a compromise is made between the real-world area covered and the grid resolution of the segmentation output. We further discuss this trade-off in section \ref{ssec:computational_challenges}. \par

\section{Network Architectures for BEV Perception} 
\label{sec:arch}

Having explored the two main tasks in the BEV space, namely, 3D object detection and \ac{BEV} segmentation, it is important to observe that the \ac{BEV} perception enables the accumulation of transformed BEV features across multiple sensors and behaves as a unified fused representation. Various algorithms can exploit the spatial and geometric reasoning of the aligned BEV features to generate more accurate outputs than directly using PV features where object positions are not realistic due to perspective distortion introduced by the onboard cameras. Consequently, the fundamental goal of \ac{BEV} perception is to map the image features in the PV to their corresponding position in the \ac{BEV} space relative to their real-world 3D locations.\par

\begin{figure}[!ht]
  \includegraphics[width=\linewidth]{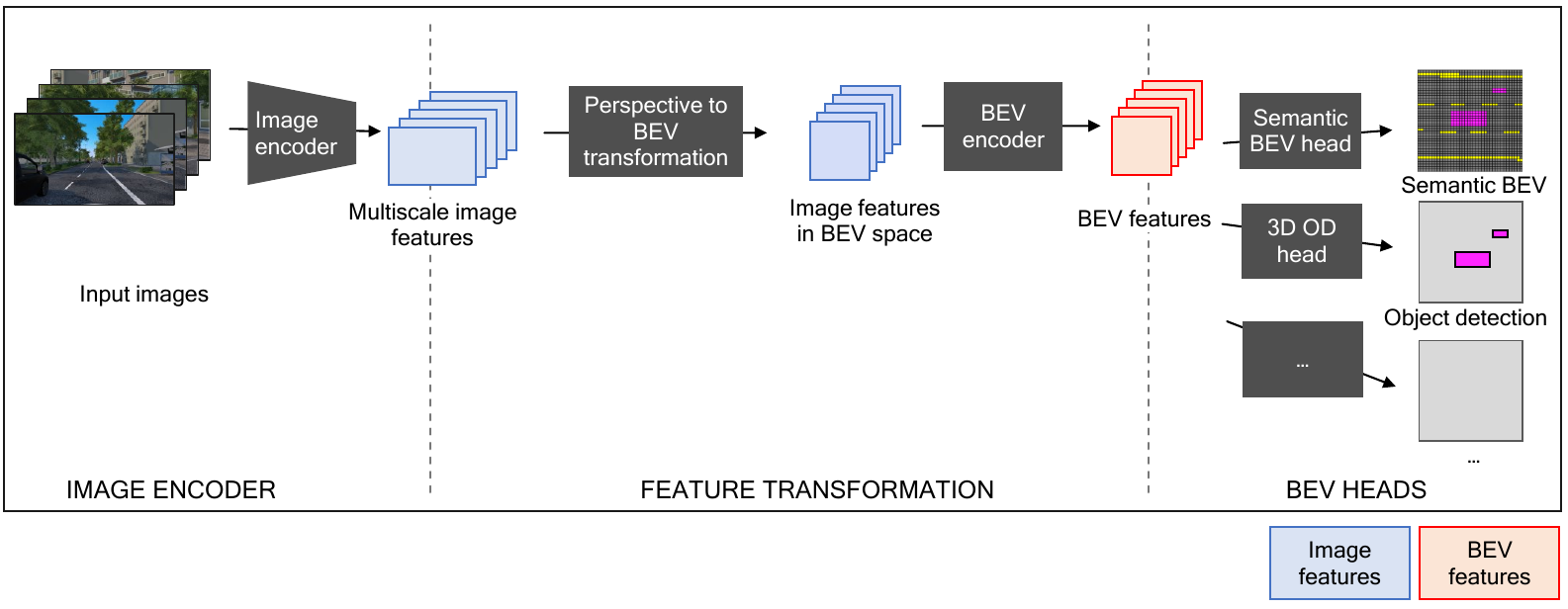}
  \caption{The architecture of \ac{BEV} transformation networks consists of typically three parts: the image encoder, the feature transformation, and the task-specific head with an optional \ac{BEV} encoder.}
  \label{fig:bev_architecture}
\end{figure}

Most recent PV-BEV transformation approaches use an end-to-end deep neural network that takes camera images and processes them to generate the desired output in the BEV space. These PV-BEV transformation networks typically comprise three main components, namely, (i) an image encoder that processes the input RGB image and generates image features in PV, (ii) a transformation module that transforms the PV features into BEV, and (iii) a task-specific head that processes BEV features and generates the task-specific output in BEV. \figref{fig:bev_architecture} illustrates the three main components of a typical PV-BEV transformation network. In this section, we explore the aforementioned network modules in detail and provide examples of the most common network architectures found in the literature for each component.\par
\subsection{Image encoder}

The image encoder, also known as a feature extractor or backbone, forms the first component of a deep neural network. Its main task is to extract features from the input image for use in the downstream modules. This module often consumes a substantial portion of computational resources compared to the whole network. The structure of the network backbone is often non-trivial. It involves engineering several hyper-parameters, such as the number of layers, the size of each layer, and connections between layers, among many others. Consequently, several works have explored various strategies to develop optimal network backbones~\cite{he2015deep, tan2020efficientnet, liu2021swin}. Several works reuse these backbone designs in BEV perception since the feature extraction step is similar in both PV and BEV perception pipelines.\par

A few of the many available network backbones in the literature have gained wide popularity in BEV perception tasks. The ResNet family of image encoders~\cite{he2015deep}, specifically ResNet-50 and ResNet-101, has been employed in both BEV-based object detection~\cite{huang2022bevdet} and BEV segmentation~\cite{roddick2020predicting}. LSS~\cite{philion2020lift} which propose the implicit estimation of depth in the image encoder to transform the features into 3D space for BEV segmentation prefer the EfficientNet~\cite{tan2020efficientnet} family of backbones due to its computational efficiency and superior performance as compared to ResNet. The same family is used by CVT~\cite{zhou2022cross} for their BEV segmentation approach, which applies cross-attention to decide for each BEV grid cell which PV features are relevant. The EfficientDet~\cite{tan2020efficientdet} image encoder extends EfficientNet with a \ac{BiFPN} layer for multi-scale feature fusion. It is used by PanopticBEV~\cite{gosala2022bird} to generate the image features for the task of BEV panoptic segmentation. Another popular class of image encoders are attention-based vision transformers such as Swin~\cite{liu2021swin}. Through their success in image classification \cite{dosovitskiy2021image}, attention based backbones are also employed for BEV perception \cite{liu2022bevfusion, huang2022bevdet}. It is important to note that switching from one image encoder to another is relatively easy. However, the model runtime and the training convergence rate heavily depend on the choice of the image encoder~\cite{philion2020lift}. This decision becomes even more relevant when multiple cameras are used since the inference time of the model can decide whether a model is deployed in the real world or not. \tabref{tab:bev_3dod} and \tabref{tab:bev_segmentation} present an overview of the image encoder used by multiple popular BEV perception approaches.\par
\subsection{\ac{BEV} transformation} 
\label{ssec: bev-transformation}

The BEV transformation module forms the second component of the PV-BEV transformation network and is responsible for transforming the image features from the PV to BEV. Existing BEV transformation modules can broadly be classified into two groups: (i) forward mapping and (ii) backward mapping. The classification is based on the BEV features generated from the PV ones. 

Forward mapping-based approaches generate the BEV features by lifting the features from the PV to the BEV using various techniques. For example, OFT~\cite{roddick2018orthographic} and M2BEV~\cite{xie2022m2bev} generate the intermediate 3D voxel grid by assuming a uniform depth distribution and accumulating the image features along the corresponding camera ray. Depth distribution is a discrete probability distribution of depth for each pixel. LSS~\cite{philion2020lift}, BEVDet~\cite{huang2022bevdet}, BEVFusion~\cite{liu2022bevfusion}, and FIERY~\cite{hu2021fiery} learn the PV-BEV mapping by first predicting the depth distribution in the PV and then using it to lift the features from the PV to the 3D space. PanopticBEV~\cite{gosala2022bird} uses a dual transformer architecture where each transformer independently attends to vertical and flat regions in the scene and transforms them from the PV to the BEV. The flat transformer employs IPM with an error correction module. In contrast, the vertical transformer implicitly learns the depth of the vertical regions to lift the corresponding features from the PV to the BEV. 

In contrast, backward mapping-based approaches learn the PV-BEV transformation by querying the value of each destination cell from the image features. These approaches are often computationally expensive since each algorithm loops through every voxel grid location to assign a value from the image features.\par

Attention-based methods recently became popular for natural language processing \cite{vaswani2017attention} and computer Vision applications \cite{dosovitskiy2021image}. Attention-based \ac{BEV} transformation represents the image-to-\ac{BEV} transformation as a translation problem from one space to another. Since, in general, a column of an image corresponds to a projected ray in \ac{BEV} space, the transformation can be interpreted as an image column to \ac{BEV} ray translation \cite{saha2022translating}. A more general attention-based transformation is to imagine the whole image to \ac{BEV} space transformation as a translation problem \cite{zhou2022cross}. They use cross-attention to determine which parts of the image space are relevant for a specific area in \ac{BEV}.\par

Geometry-based transformation approaches use the camera parameters like position, orientation, or focal length, which are given by the extrinsic and intrinsic camera parameters and are known for cameras used in the autonomous driving context. Based on those parameters, one can calculate the 3D position of the image pixel up
to the ambiguous depth of the pixel. Geometry-based transformation approaches use these geometrical constraints to project image features into \ac{BEV} feature space. Early works like OFT project the image features into 3D space, assuming a uniform distribution for the depth distribution \cite{roddick2018orthographic}. The seminal
work from LSS uses the image encoder backbone to estimate the depth distribution for each image feature additionally. It uses this depth distribution to project the image feature into 3D space.
As of now, geometry-based architectures show the most promising overall results, so we limit further architectural discussion to this group of works and elaborate only on those in the following more detailed subsection.\par

Geometry-based \ac{BEV} transformation approaches show great performance since they model the real world using a 3D model that is condensed in a subsequent step. Recent architectures use a four-step transformation approach~\cite{philion2020lift, huang2022bevdet}: First, a primarily \ac{CNN}-based image encoder generates image features from the input images. Second, the view transformation module transforms the image features from the 2D perspective space into the \ac{BEV} space. Then a \ac{BEV} encoder processes the features using a convolutional network in the \ac{BEV} space. Finally, the encoded \ac{BEV} features are fed into one or several task-specific heads, e.g., for semantic segmentation, 3D object detection, or both.\par
\begin{figure}[!ht]
  \includegraphics[width=\linewidth]{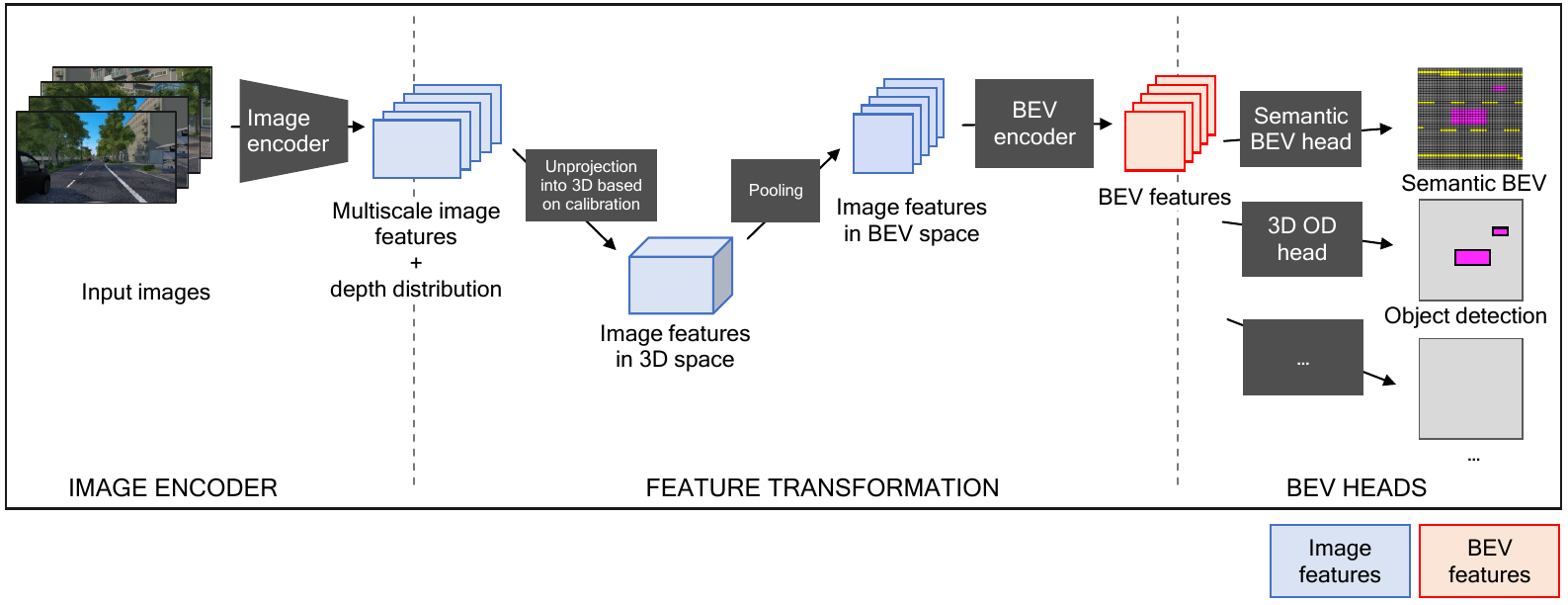}
  \caption{\ac{BEV} architecture of LSS' two-step transformation approach: They take the image features and the predicted depth distribution and project them into the 3D space. Then they pool the 3D space in the height dimension to obtain the \ac{BEV} representation.}
  \label{fig:lss_architecture}
\end{figure}
\begin{figure}[!ht]
  \includegraphics[width=12cm]{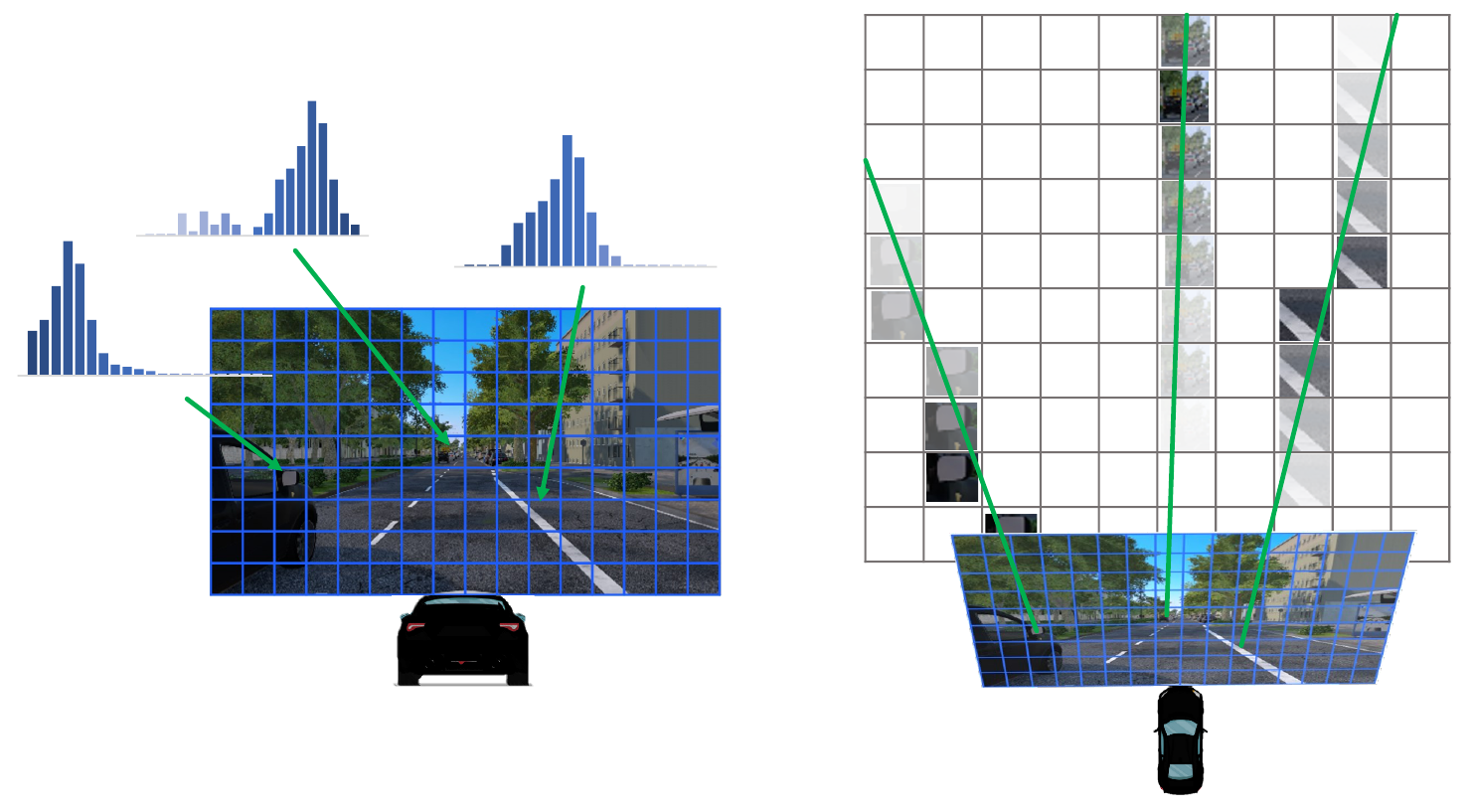}
  \caption{The network backbone estimates the features and, in addition, the depth distribution for each feature. As a second step, the image features are projected into the 3D space based on the depth distribution and the camera parameters.}
  \label{fig:lss_drawing}
\end{figure}
\figref{fig:lss_architecture} shows the transformation approach from LSS \cite{philion2020lift}, that is widely used by subsequent works \cite{liu2022bevfusion}, \cite{huang2022bevdet}. The backbone image encoder is shared between all camera images and outputs the image space features and a predicted depth distribution for each image feature. This depth distribution is used to lift the image features into 3D space, as sketched in \figref{fig:lss_drawing}. Each image feature is projected into the 3D space along a ray. The ray direction matches the direction where the light ray hits the camera and is determined by the camera intrinsics. The depth estimation part of the network backbone determines how intense the features are projected to each distance step. Each projected 3D feature is assigned to the corresponding 2D \ac{BEV} grid cell in the splat step. Features that are projected outside the valid grid area are dismissed. After this lift-splat step, the features exist in \ac{BEV} space and are processed by the \ac{BEV} encoder. The encoded \ac{BEV} features are fed into one or several task-specific heads to provide the desired output.\par
\begin{table}[!ht]
\centering
\caption{Different BEV segmentation approaches.}
\label{tab:bev_segmentation}
\resizebox{\textwidth}{!}{%
\begin{tabular}{@{}l|lllll@{}}
\toprule
\textbf{Name}                      & \textbf{Datasets}                                                                 & \textbf{\begin{tabular}[t]{@{}l@{}}Multi\\      Camera\end{tabular}} & \textbf{\begin{tabular}[t]{@{}l@{}}Input \\      Resolution\end{tabular}}                                                        & \textbf{Encoder} & \textbf{\begin{tabular}[t]{@{}l@{}}Grid Dimension /\\      Cell Size\end{tabular}}                                                                                                                                                          \\ \midrule
VED   \cite{lu2019monocular}       & \begin{tabular}[t]{@{}l@{}}Cityscapes, \\      KITTI\end{tabular}                 & no                                                                   & 256 × 512                                                             & VGG-16           & 64x64                                                                                                                                                                       \\
VPN   \cite{pan2019cross}          & \begin{tabular}[t]{@{}l@{}}House3D,  \\      CARLA, \\      nuScenes\end{tabular} & yes                                                                  &                                                                       & ResNet-18        &                                                                                                                                                                             \\
PON   \cite{roddick2020predicting} & \begin{tabular}[t]{@{}l@{}}nuScenes, \\      Argoverse\end{tabular}               & no                                                                   &                                                                       & ResNet-50        & \begin{tabular}[t]{@{}l@{}}200x200 /  0.25m\end{tabular}                                                                          \\
LSS   \cite{philion2020lift}       & \begin{tabular}[t]{@{}l@{}}nuScenes, \\      Lyft\end{tabular}                    & yes                                                                  & 128 × 352                                                             & EfficientNet-B0  & \begin{tabular}[t]{@{}l@{}}200x200 /  0.5m\end{tabular}                                                                         \\
TIIM   \cite{saha2022translating}  & \begin{tabular}[t]{@{}l@{}}nuScenes, \\      Argoverse, \\      Lyft\end{tabular} & no                                                                   &                                                                       & ResNet-50        & \begin{tabular}[t]{@{}l@{}}200x200 /  0.25m\end{tabular}                                                                          \\
PanSeg   \cite{gosala2022bird}     & \begin{tabular}[t]{@{}l@{}}KITTI-360, \\      NuScences\end{tabular}              & no                                                                   & \begin{tabular}[t]{@{}l@{}}768 × 1408, \\      448 × 768\end{tabular} & EfficientDet     &                                                                                                                                                                             \\
BEVFormer   \cite{li2022bevformer} & \begin{tabular}[t]{@{}l@{}}nuScenes, \\      Waymo\end{tabular}                   & yes                                                                  & 900 × 1600                                                            & ResNet101-DCN    & \begin{tabular}[t]{@{}l@{}}200x200 /  0.512m\end{tabular} \\
GitNet   \cite{gong2022gitnet}     & \begin{tabular}[t]{@{}l@{}}nuScenes, \\      Argoverse\end{tabular}               & no                                                                   &                                                                       & ResNet-50        & \begin{tabular}[t]{@{}l@{}}200x200 /  0.25m\end{tabular}                                                                          \\
M2BEV   \cite{xie2022m2bev}            & nuScenes                                                                          & yes                                                                  & 900 × 1600                                                            & ResNeXt-101      & \begin{tabular}[t]{@{}l@{}}200x200 /  0.5m\end{tabular}                                                                         \\
CVT   \cite{zhou2022cross}         & \begin{tabular}[t]{@{}l@{}}nuScenes, \\      Argoverse\end{tabular}               & yes                                                                  &                                                                       & EfficientNet-B4  & \begin{tabular}[t]{@{}l@{}}200x200 /  0.5m\end{tabular}                                                                         \\ 
\bottomrule
\end{tabular}%
}
\end{table}
\begin{table}[!ht]
\centering
\caption{Different 3D object detection approaches.}
\label{tab:bev_3dod}
\resizebox{\textwidth}{!}{%
\begin{tabular}[t]{@{}l|lllll@{}}
\toprule
\textbf{Name}                         & \textbf{Datasets}                                               & \textbf{\begin{tabular}[t]{@{}l@{}}Multi \\      Camera\end{tabular}} & \textbf{\begin{tabular}[t]{@{}l@{}}Input \\      Resolution\end{tabular}}                                                           & \textbf{Encoder}                                                                   &
\textbf{\begin{tabular}[t]{@{}l@{}}Grid Dimension /\\      Cell Size\end{tabular}}  
                             \\ \midrule
OFT   \cite{roddick2018orthographic}  & KITTI                                                           & no                                                                    & -                                                                        & ResNet-18                                                                          & \begin{tabular}[t]{@{}l@{}}160x160 / 0.5m\end{tabular}   \\
CaDDN   \cite{reading2021categorical} & \begin{tabular}[t]{@{}l@{}}KITTI, \\      Waymo\end{tabular}    & no                                                                    & 832 × 1248                                                               & ResNet-101                                                                         & \begin{tabular}[t]{@{}l@{}}320x336 / 0.16m\end{tabular}   \\
BEVFormer   \cite{li2022bevformer}    & \begin{tabular}[t]{@{}l@{}}nuScenes, \\      Waymo\end{tabular} & yes                                                                   & 900 × 1600                                                               & ResNet-101                                                                         & \begin{tabular}[t]{@{}l@{}}200x200 / 0.512m\end{tabular} \\
BEVDet   \cite{huang2022bevdet}       & nuScenes                                                        & yes                                                                   & 256 × 704                                                                & \begin{tabular}[t]{@{}l@{}}ResNet-50, \\      ResNet-101,\\      Swin\end{tabular} & \begin{tabular}[t]{@{}l@{}}128x128 / 0.8m\end{tabular}   \\
BEVDet4D\cite{huang2022bevdet4d}      & nuScenes                                                        & yes                                                                   & \begin{tabular}[t]{@{}l@{}}256 × 704 × 2 \\      (temporal)\end{tabular} & ResNet-50                                                                          & \begin{tabular}[t]{@{}l@{}}128x128 / 0.8m\end{tabular}   \\
M2BEV   \cite{xie2022m2bev}               & nuScenes                                                        & yes                                                                   & 900 × 1600                                                               & ResNeXt-101                                                                        & \begin{tabular}[t]{@{}l@{}}200x200 / 0.5m\end{tabular}   \\
BEVDepth   \cite{li2022bevdepth}      & nuScenes                                                        & yes                                                                   & \begin{tabular}[t]{@{}l@{}}256 × 704, \\      512 × 1408\end{tabular}    & ResNet-50                                                                          &                                                                           \\ \bottomrule
\end{tabular}%
}
\end{table}
\subsection{Task-specific head}

The network head forms the last step of a typical neural network pipeline and is responsible for generating the required task-specific output. The BEV head accepts the transformed BEV features from the PV-BEV transformation module as input. It generates the 3D bounding box regressions for the 3D object detection task and either BEV semantic, instance or panoptic maps in case of the BEV segmentation task. The task-specific heads in the BEV representation have largely remained unchanged from the existing PV-based and 3D LiDAR-based approaches, apart from minor modifications to adapt them to the characteristics of the novel view.\par

Focusing on the task of 3D object detection, no consensus is found in the literature, and different authors employ different object detection heads in their proposed frameworks. OFT~\cite{roddick2018orthographic} takes inspiration from DenseBox~\cite{huang2015densebox} and predicts a confidence map containing the coarse locations of objects in the scene. This output is augmented with three other heads that predict the position offsets to improve object localization and the size and orientation of the 3D bounding boxes. CaDNN~\cite{reading2021categorical} and M2BEV~\cite{xie2022m2bev} employ the popular 3D LiDAR detection head, PointPillars~\cite{lang2019pointpillars}, as their object detection head. The former employs the vanilla PointPillar architecture. The latter augments it with a dynamic box assignment strategy that learns to dynamically assign 3D anchors to ground truth boxes instead of using a fixed IoU threshold. BEVFormer~\cite{li2022bevformer} adapts the 2D detector deformable DETR~\cite{wang2021detr3d} to predict 3D bounding boxes and uses it as part of its 3D object detection pipeline. Lastly, BEVDet~\cite{huang2022bevdet} and BEVDet4d~\cite{huang2022bevdet4d} use the first stage of CenterPoint~\cite{yin2021center} augmented with a novel scaleNMS strategy as their object detection head. ScaleNMS scales the bounding box predictions of each object based on its class before applying NMS to ensure that the NMS algorithm is effective even in BEV space, where the overlap between the bounding boxes is close to zero.\par

Similar to 3D object detection, no agreement is found in the literature regarding the architecture of the BEV segmentation head. VED~\cite{lu2019monocular} uses six upsampling modules, each consisting of a transpose convolutional layer followed by two convolutional layers. In contrast, VPN~\cite{pan2019cross} uses a pyramid pooling module~\cite{zhao2017pyramid} for generating the BEV semantic predictions from the BEV features. PON \cite{roddick2020predicting} employs a stack of 8 residual blocks and a transpose convolutional layer to generate the segmentation map. At the same time, LSS~\cite{philion2020lift} uses the PointPillar~\cite{lang2019pointpillars} architecture to flatten the intermediate 3D features into BEV and generate the semantic maps in the BEV space. TIIM~\cite{saha2022translating} employs multiple ResNet blocks with Deep Layer Aggregation~\cite{yu2018deep} to generate the BEV maps. Lastly, PanopticBEV~\cite{gosala2022bird} uses DPC and LSFE modules along with depthwise separable convolutions as proposed in EfficientPS~\cite{mohan2021efficientps} to generate the required BEV semantic map from the intermediate features. Lastly, M2BEV~\cite{xie2022m2bev} uses a very simple 5-layer convolutional block with 3x3 and 1x1 kernels to generate the BEV segmentation predictions.\par
\section{Datasets and Metrics for BEV Perception} 
\label{sec:datasets}

Publicly available datasets are significant for research because they reduce the barrier to entry and prevent the need for resource and time-intensive data collection and annotation. In addition, public datasets make different approaches comparable with each other since they can be tested using the same data constraints. Considering that \ac{BEV} perception is a relatively new research area, there are, to our knowledge, no datasets designed explicitly
for \ac{BEV} perception. However, many datasets for autonomous driving contain 3D bounding box annotations and semantic maps that can be used for \ac{BEV} perception\par
\subsection{Datasets}

Most works in the \ac{BEV} field \cite{ma2022vision} use the nuScenes dataset for their research \cite{caesar2020nuscenes}. The main reason is that it provides full 360\,° camera field of view and accompanying HD map with Lidar to provide BEV ground truth. The nuScenes dataset comprises six pinhole cameras that yield a combined 360\,° field of view around the vehicle, in addition to one LiDAR sensor and five Radars. They annotate 3D bounding boxes for 23 classes, such as vehicles and pedestrians. Additionally, they provide a detailed HD map with eleven semantic classes that can be used for \ac{BEV} segmentation. nuScenes includes 1000 scenes from Boston and Singapore with each 20\,s length and provides a strong baseline for \ac{BEV} perception data. Moreover, the Waymo Open Dataset is like nuScenes but includes five LiDAR sensors and, therefore, better availability of depth information~\cite{sun2020scalability}. It contains 1150 scenes of each 20\,s, but with more annotated frames, which is why it is larger than the nuScenes. The Waymo Open Dataset cameras all face to the front or the sides; this is why they do not cover the full 360\,° for \ac{BEV} perception, but only around 240°. Therefore, it is unsuitable for 360\,° \ac{BEV} perception. Furthermore, some works like TIIM~\cite{saha2022translating} and
GitNet \cite{gong2022gitnet} additionally report their performance on the Argoverse dataset~\cite{chang2019argoverse}. In terms of labels, Argoverse provides both 3D bounding boxes and semantic maps and provides even the benefits of having, in addition to the 360\,° field of view, a pair of front-facing stereo cameras. Argoverse contains fewer scenes than nuScenes or the Waymo Open dataset, which is why it is less used.
The KITTI-360 dataset provides 3D bounding boxes and semantic map information for a pair of front-viewing stereo cameras and two fisheye cameras. Fisheye cameras are used by many \acp{OEM} for their \ac{ADAS} since fisheye cameras are often utilized for parking visualization. Table \ref{tab:datasets} gives a more detailed overview of the various public datasets that can be used for \ac{BEV} perception.\par
\begin{table}[!ht]
\centering
\caption{Autonomous driving datasets can be used for \ac{BEV} perception.\\\hspace{\textwidth} \textsuperscript{*} p: pinhole, s: stereo pair, f: fisheye}
\label{tab:datasets}
\resizebox{\textwidth}{!}{%
\begin{tabular}{@{}lllllllllll@{}}
\toprule
\textbf{Dataset} & \textbf{Location} & \textbf{Year} & \textbf{Scenes} & \textbf{Train} & \textbf{Val} & \textbf{Test} & \textbf{3D Boxes} & \textbf{Maps} & \textbf{Cameras *} & \textbf{Lidars} \\ 
\midrule
KITTI      & Karlsruhe (GER)   & 2012 & 22    & 7k   & -   & 7k   & 200k    & No  & 2 s       & 1 \\
nuScenes   & Boston, Singapore & 2019 & 1000  & 28k  & 6k  & 6k   & 1,400k  & Yes & 6 p       & 1 \\
Waymo Open & 3x USA            & 2019 & 1150  & 170k & 30k & 30k  & 12,000k & No  & 5 p       & 5 \\
Argoverse  & 2x USA            & 2019 & 113   & 39k  & 15k & 12k  & 993k    & Yes & 7 p + 1 s & 2 \\
KITTI-360  & Karlsruhe (GER)   & 2021 & 11    & 78k  & -   & -    & 68k     & Yes & 1 s + 2 f & 2 \\
Argoverse  2 & 6x USA          & 2021 & 1000  & -    & -   & -    & -       & Yes & 7 p + 1 s & 2 \\ 
\bottomrule
\end{tabular}%
}
\end{table}
In addition to the mentioned public datasets, it can be useful to use a synthetic dataset generated by a 3D simulation tool like CARLA \cite{dosovitskiy2017carla}. Using 3D simulation, generating the desired information accurately with minimal labeling errors is simple. In the simulation, a camera can be mounted above the car and follow the car like a drone to record the scene for generating the \ac{BEV} perception labels. The synthetic \ac{BEV} camera has the advantage that it does not introduce measurement errors and that most simulation tools can output both per-pixel depth information and per-pixel semantic information, which leads to accurate labels for \ac{BEV} segmentation. In addition, simulators can output objects as instances with precise position and size, which is more accurate than the labels for real-world datasets generated by LiDAR data which still have measurement errors. Training models with synthetic data come with a downside called the reality gap~\cite{jakobi1995noise}. Since the simulation is only a simplified model of reality, the simulated data is generally less complex and has different statistical properties than the real-world data. This gap between simulation and reality often results in substantial degradation when testing a simulation-trained neural network for real-world problems and requires countermeasures.\par
\subsection{Metrics}

\begin{wrapfigure}{r}{0.3\textwidth}
  \begin{center}
    \includegraphics[width=0.3\textwidth]{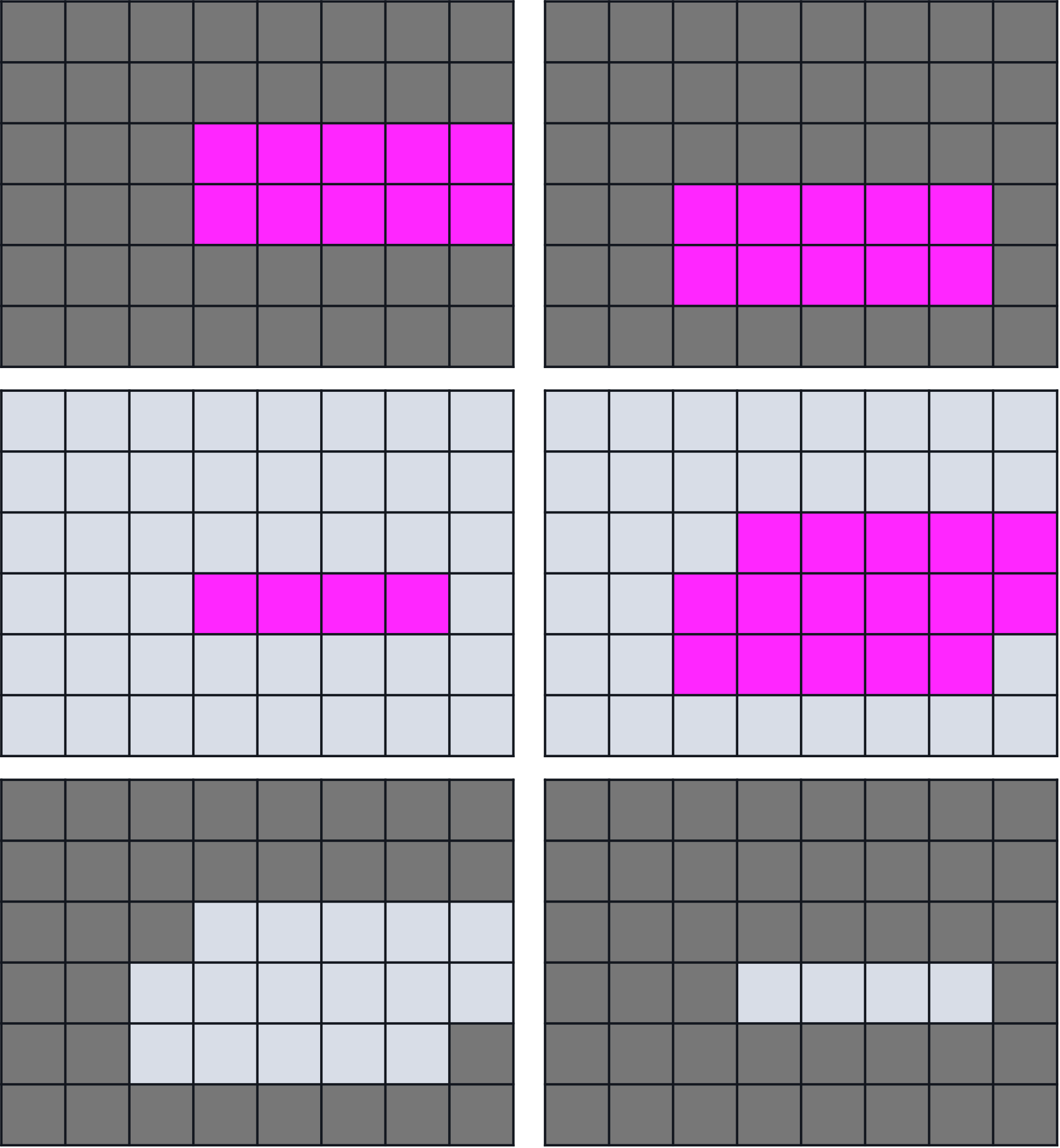}
    \caption{Top: GT and prediction of a vehicle. Middle: Intersection and union for the class vehicle. $IoU = 4/16 = 0.25$. Bottom: Intersection and union for the background class. $IoU = 32/44 = 0.73$}
    \label{fig:iou}
  \end{center}
\end{wrapfigure}

It is crucial to have suitable metrics to evaluate different methodologies for BEV perception and to measure progress during training. The standard metric for \ac{BEV} segmentation is \ac{IoU}, which is also standard for image segmentation tasks. The \ac{IoU} score for each class is defined in equation \ref{eq:iou} where $y$ denotes the ground truth and $\hat{y}$ denotes the predicted output.\par
\begin{equation} 
    IoU_c(y, \hat{y}) = \frac{ \left | y_c\cap \hat{y}_c \right | }{\left | y_c \cup \hat{y}_c \right |}
  \label{eq:iou}
\end{equation}
Naturally, classes with a more significant representation in the real world, both in size and frequency, such as streets, vegetation, and buildings, achieve a higher IoU w.r.t instances like pedestrians. Figure~\figref{fig:iou} visualizes this behavior, where the vehicle class scores lower than the background class. The mean across \acp{IoU}s of all classes is the \ac{mIoU} metric and is often used to compare semantic segmentation approaches.\par
For 3D object detection, the primary metric is \ac{mAP}, which is explained in the following. The \ac{mAP} metric compares detected and ground truth objects, and its calculation requires several steps. The first computation step evaluates if an object is detected or not. The KITTI dataset \cite{geiger2012are} proposes to use the bounding box overlap or \ac{IoU} per object to determine a correct detection. If a detected object reaches the IoU threshold of $50\%$, it counts as a \ac{TP}, otherwise as a \ac{FN}. Falsely detected objects that do not have an \ac{IoU} with the ground truth, which lays above the threshold, count as \ac{FP}. The nuScenes dataset \cite{caesar2020nuscenes} compute the correctness of a detection differently. They use the 2D center distance on the ground plane as a threshold, decoupling the detection rate from the object size. If detection and ground truth are within a distance threshold, the detection counts as a \ac{TP}, otherwise as a \ac{FN}. Based on \ac{TP}, \ac{FP} and \ac{FN}, the intermediate measures precision and recall can be calculated according to formulas \ref{eq:precision-recall}. Precision states how accurate a detector is when detecting a class, whereas recall states the percentage of all objects of a class the detector can detect. A detector can achieve a high precision score when predicting only distinct detections and missing less clear ones. Furthermore, a detector can achieve a high recall score when predicting many objects, even unclear ones. This contrast shows that each metric alone is not meaningful, but combining both is an expressive metric.\par
\begin{equation} 
\label{eq:precision-recall}
\text{precision} = \frac{TP}{TP + FP} / / \text{recall} = \frac{TP}{TP + FN}
\end{equation}
Based on precision and recall, a precision-recall curve can be constructed by computing the precision value obtained when assuming a fixed recall value. The final object detection metric \ac{AP} can be calculated based on the precision-recall curve by computing the \ac{AUC} by integrating the precision $p$ over the recall $r$ as stated in \eqref{eq:AP}. 
\begin{equation} 
\label{eq:AP}
AP = \int_{0}^{1} p(r) dr
\end{equation}
\begin{wrapfigure}{r}{0.4\textwidth}
  \begin{center}
    \includegraphics[width=0.4\textwidth]{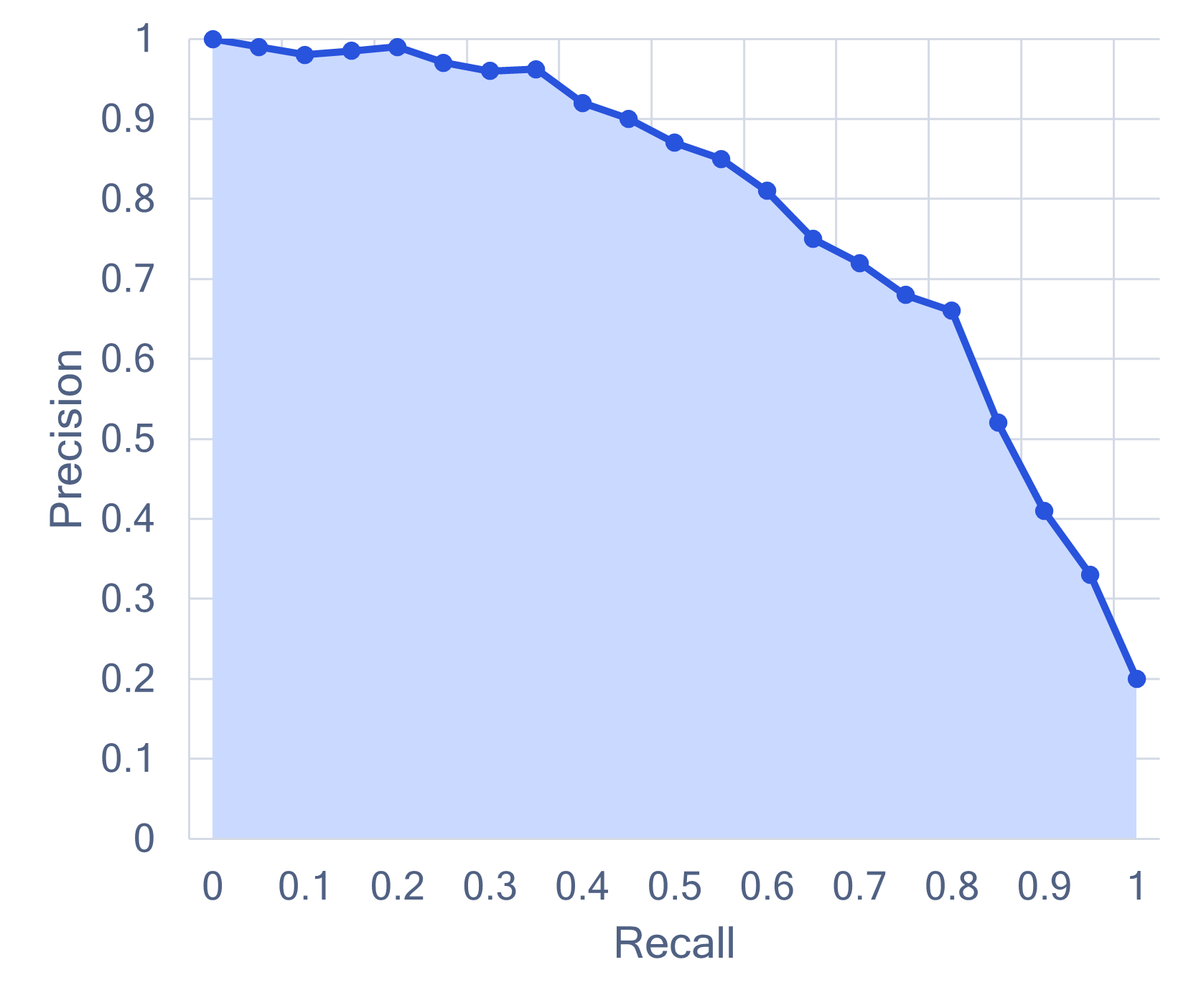}
    \caption{The PR curve (dark blue) and \ac{AUC} measure (light blue) for an example detector}
    \label{fig:precision_recall_curve}
  \end{center}
\end{wrapfigure}
Ordinarily, \ac{mAP} is used to measure the final performance. \ac{mAP} can be calculated from \ac{AP} by averaging the dataset dependent on the different classes, subclasses, or scenario difficulty levels. \ac{mAP} only focuses on the position and the overlap of a detected object with the corresponding ground truth but leaves substantial other performance indicators aside. Therefore, Caesar et al. propose the \ac{NDS} in their work. NDS is a weighted score that is composed of $50\%$ mAP metric. The other $50\%$ of the score takes box location, size, orientation, attributes, and velocity into consideration \cite{caesar2020nuscenes}.\par
\section{Challenges and Open Problems in \ac{BEV} perception} 
\label{sec:challenges}

Literature on \ac{BEV} perception demonstrates good performance on various tasks such as \ac{BEV} segmentation or 3D object detection. As the success of these algorithms is relatively new, \ac{BEV} perception is currently in a transitioning process into the industry. There remain unsolved challenges that need to be resolved for \ac{BEV} perception to become an overarching framework for the usage in commercial deployment projects where a whole fleet of vehicles is involved. This section discusses open topics and encourages new research in those areas.\par
\subsection{Extension to fusion and downstream tasks}

The autonomous driving stack from sensor, perception, situation analysis, and path planning to actuation and control is quite complex. This subsection elaborates on how \ac{BEV} perception can be extended and used for downstream tasks from perception.\par

One significant advantage of the representation of \ac{BEV} appears in the multi-modal sensor fusion or step across Cameras, LiDARs and Radars. LiDAR and Radar provide 3D measurements that can easily be converted into \ac{BEV} space. BEVFusion uses camera and LiDAR features and feeds both together in a shared \ac{BEV} encoder~\cite{liu2022bevfusion} and achieves a state-of-the-art. Another work shows that concatenating camera features in \ac{BEV} space with Radar detections boosts the performance for \ac{BEV} segmentation when no LiDAR sensor is available~\cite{harley2022simple}. In addition, a network can learn not only about what it sees but also about what it does not see. It can be achieved by training the network with an additional parameter for occlusion probability~\cite{reiher2020sim2real}, which can help handle situations with hidden road users.\par

Besides the apparent advantage of multi-modal sensor fusion, \ac{BEV} space representation is also well suited for several other tasks for situation analysis and planning. The situation analysis (future trajectory prediction for actors, collision detection and avoidance, and other tasks) becomes even more effective when considering multiple \ac{BEV} frames instead of just single ones. Using multiple frames enables temporal fusion, increasing the accuracy because it is less likely that the detector misses the same object for several consecutive frames. In addition, temporal fusion can improve scenes where essential aspects of a set are occluded for some frames but visible during others. Many works report performance increases for \ac{BEV} segmentation and 3D object detection when enabling temporal fusion \cite{saha2021enabling}, \cite{saha2022translating}, \cite{huang2022bevdet4d}. However, not only fusion profits when evaluating the situation based on multiple frames. Optical flow in \ac{BEV} space can be easily computed and has the advantage that optical flow directly corresponds to velocities in the real world. Therefore, the velocity of the ego vehicle and the other road users can be calculated accurately based on the optical flow in \ac{BEV}. We want to mention the benefits of mapping and localization to complete the advantages of situation analysis. Maps can be generated from the \ac{BEV} output by compensating for the ego-motion and combining all parts of the \ac{BEV} that are usually fixed and do not move between time steps. Further, a vehicle can localize on HD maps based on BEV outputs. HD maps enhance the scene understanding as they provide very accurate environment information.\par

For trajectory planning, it is not only important to know what the current situation is, but we also need to reason about the plausible future paths of other vehicles. FIERY~\cite{hu2021fiery} proposes a model in \ac{BEV} space, where they predict probabilistic future trajectories of other road users. This information can improve trajectory prediction. Generally, trajectory planning is done by evaluating the current situation using conventional programming methods and then proposing a path based on the result. A neural network can perform trajectory planning for the ego vehicle in \ac{BEV} space by proposing a trajectory based on trajectories from a training dataset. LSS, for example, proposes a method to predict potential trajectories in \ac{BEV} space \cite{philion2020lift}.\par
\subsection{Perception challenges}

\ac{BEV} perception with monocular cameras being the only input modality is a challenging task thus far. This is due to the inability of monocular cameras to provide depth information, which causes inaccuracies in the \ac{BEV} Transformation illustrated in Section~\ref{ssec: bev-transformation}. While some applications, such as 3D object detection for vehicles, achieve reliable results, other tasks, such as Occupancy Flow prediction, are more ambitious. The ambiguity in depth particularly affects small objects in the \ac{BEV} space. For instance, pedestrians are hard to annotate accurately by the network. This problem worsens in faraway regions, as the errors in transforming distant regions from the Perspective View into the orthographic \ac{BEV} space are substantially higher.\par
\subsection{Practical challenges}

Many changes are required when transforming a computer vision architecture from an image-space-based architecture toward a \ac{BEV}-space-based architecture. One cannot just replace the PV encoder architectures for segmentation and detection and keep the other parts in a BEV framework unmodified. The perception model itself would require extensive modifications. Moving towards \ac{BEV} perception also means moving towards a more unified perception architecture where the information from the different cameras is not combined after the different task-specific heads but instead combined at feature level in \ac{BEV} space. This alignment is learned by the network end-to-end, simplifies the post-processing, and automates the effort to consolidate the information from different cameras into a unified representation. On the other hand, having a data-driven approach for the feature consolidation requires a new set of labels in \ac{BEV} space to train the new end-to-end architecture. In practice, this means that a large-scale \ac{BEV} dataset needs to be created, which requires labeling tools to annotate in \ac{BEV} space.\par
\subsection{Computational challenges}\label{ssec:computational_challenges}

There are computational challenges to using deep learning: during training on the server side and inference on the edge side. Since computations in 3D space, especially 3D convolutions, require many operations and increase the FLOPS for training and inference, it is beneficial to avoid computations in the 3D space as much as possible. The other two main factors determining the computational load are input and output resolution. Some works use an input resolution as low as 0.05\,\ac{MP} per image \cite{philion2020lift}, which has the advantage that the image encoder can be small and does not require many resources. The computational effort is much larger when switching to images with resolutions around 1\,\ac{MP} or above. It becomes challenging to compute this high-resolution camera without adaptions to optimize performance. In industry, there is a trend towards an 8\,\ac{MP} front camera, which makes a compute-optimized architecture for \ac{BEV} perception essential.\par

Another challenge is the design of a suitable output format. When predicting \ac{BEV} segmentation, there is a trade-off between grid resolution and detection range. If one wants to detect objects up to 200\,m around the car, which is especially required at high speed, and in addition needs accuracy of 0.1\,m to sense lane markings and pedestrians, the grid output would be $200\,m \times 2 / 0.1\,m = 4000\,pixel$ per side of the \ac{BEV} grid. This grid size would not only be too large to compute but also require too many parameters that would tend to overfit and, therefore, not feasible. On the other hand, the suggested 0.1\,m accuracy is mainly required for the near range, for example, 20\,m around the car, as precise localization of pedestrians and lane markings is less important further away from the ego vehicle. Therefore, the far range could be sampled with a lower resolution and still provide the relevant information. This reduction in spatial resolution with the distance from the ego vehicle is similar to how we as humans perceive the environment.\par

There are several potential solutions. One could think about a non-uniform grid cell size. The grid cells close to the ego-vehicle would be small and have a high resolution, while far away cells would have a lower resolution and larger grid cell size. While this is mathematically possible, a non-uniform \ac{BEV} grid comes with a crucial downside as it is no longer translation equivariant, which makes it harder to create and
complicates further processing. A better option is probably to create a near-field and a far-field \ac{BEV} grid. A possible configuration could be a near-field grid with $0.125$\,m grid cell size and $25$\,m range in each direction, which would require a total of $400\times400$ grid cells and would be suitable for near-range reasoning, mainly during low-speed maneuvering and for parking situations. On the other hand, the far-field
grid could be resolved with $0.5$\,m grid cell size and a range of $200$\,m to the front and rear and $100$\,m to the sides, resulting in a grid of $800\times 400$\ pixels. Using this dual grid setup would result in a total output resolution of $0.48$\,\ac{MP}, which is still relatively high compared to the current state-of-the-art resolution in research, which is $200\times200$, pixels or $0.04$\,\ac{MP}. This increase in output resolution requires still significantly more computing and a rich dataset that comprises near-range and far-range sensors to train the larger network.\par
\subsection{Geometric challenges}

Since \ac{BEV} perception depicts the real-world occupancy information, it needs to model the real world accurately to achieve reliable results. This modeling process is realized as a combination of the \ac{BEV} feature transformation step and the learning within the model weights. A scene with some vehicles on a flat street can be modeled relatively easily when using the camera parameters to project the features. More complex scenarios encompass, e.g., uneven road surfaces, roads with inclines, up and down winding roads, or changing extrinsic camera parameters through roll or pitch angles of the vehicle suspension. When not handled
geometrically by the transformation approach or by adapted camera parameters, the \ac{BEV} perception is susceptible to those situations since the large variety of situations cannot be all part of the training data. Therefore, a pure learning approach cannot handle those situations well. Intuitively, it makes sense to compensate for the camera rotation parameters by the roll and pitch angles of the suspension when they are precisely available. To tackle the different road geometries, an additional road surface estimation that predicts inclines might prove helpful. The transformation model would be required to input this road surface estimation and use the information efficiently to ease the modeling part for the network weights and to reduce the learning effort.\par
\subsection{Limitations}

The \ac{BEV} representation is well suited for many tasks around autonomous driving, but the simplification from 3D to \ac{BEV} comes at the cost of missing height information, which is still required for some tasks. Information like the over-drivable and under-drivable properties of objects can be provided as additional object parameters and are not a direct problem for the \ac{BEV} representation. However, for information like traffic signs, traffic lights, and pedestrian poses, additional object parameters are very inefficient in \ac{BEV} space, which is why this information should be handled differently.\par

Existing methods for the mentioned tasks work well in image space, which is why the tasks should ideally still be handled in the image space. It can be part of a coherent architecture, with heads for traffic signs, traffic lights, and poses after the image encoders to perform the image space tasks. In addition, the encoded image features are fed into the \ac{BEV} transformation module to enable \ac{BEV} perception.\par
\section{Conclusion}

\ac{BEV} based representation shows superior performance for 3D semantic segmentation and 3D object detection, especially in accurate 3D localization, based on the reasoning in 3D space. As \ac{BEV} perception directly learns to combine features of multiple cameras into a coherent representation, it avoids the manual effort of combining outputs from different cameras. In addition, \ac{BEV} perception shows the potential to perform multi-sensor fusion between camera, LiDAR, and Radar. This shows why \ac{BEV} perception can be a foundation for the overall 3D perception of autonomous vehicles. In this chapter, we highlighted the different architectures to implement \ac{BEV} perception and further elaborated on potential downstream extensions, challenges, and corresponding future research areas. The main motivation of \ac{BEV} perception is map the pixel-space 2D input to 3D output which is necessary for downstream tasks in automated driving which operate in 3D. This type of cross-view perception where the input and output domains are different is not a typical problem in computer vision. The tools used in standard dense prediction tasks such as data augmentation and hard negative mining need to be adapted. This is a critical step to improve the maturity of this rapidly progressing research area. However, the fundamental challenge is to learn an optimal feature embedding in 3D using 2D pixels which have depth ambiguity. \par
\section*{Acknowledgement}

The synthetic data used in this book chapter were generated using a commercial synthetic engine www.cognata.com. The authors thank Ahmed Sadek and Ravi Kiran for the detailed review and feedback.\par
\cleardoublepage
\bibliographystyle{plain}
\bibliography{bibtex_example}

\begin{thebibliography}{10}

\bibitem{borse2023x}
Shubhankar Borse, Marvin Klingner, Varun~Ravi Kumar, Hong Cai, et~al.
\newblock {X-Align: Cross-Modal Cross-View Alignment for Bird's-Eye-View Segmentation}.
\newblock In {\em Proceedings of the IEEE/CVF Winter Conference on Applications of Computer Vision}, pages 3287--3297, 2023.

\bibitem{caesar2020nuscenes}
Holger Caesar, Varun Bankiti, Alex~H Lang, Sourabh Vora, Venice~Erin Liong, Qiang Xu, Anush Krishnan, Yu~Pan, Giancarlo Baldan, and Oscar Beijbom.
\newblock {nuscenes: A multimodal dataset for autonomous driving}.
\newblock In {\em Proceedings of the IEEE/CVF conference on computer vision and pattern recognition}, pages 11621--11631, 2020.

\bibitem{chang2019argoverse}
Ming-Fang Chang, John Lambert, Patsorn Sangkloy, Jagjeet Singh, Slawomir Bak, Andrew Hartnett, De~Wang, Peter Carr, Simon Lucey, Deva Ramanan, et~al.
\newblock {Argoverse: 3d tracking and forecasting with rich maps}.
\newblock In {\em Proceedings of the IEEE/CVF Conference on Computer Vision and Pattern Recognition}, pages 8748--8757, 2019.

\bibitem{dhananjaya2021weather}
Mahesh~M Dhananjaya, Varun~Ravi Kumar, and Senthil Yogamani.
\newblock {Weather and light level classification for autonomous driving: Dataset, baseline and active learning}.
\newblock In {\em 2021 IEEE International Intelligent Transportation Systems Conference (ITSC)}, pages 2816--2821. IEEE, 2021.

\bibitem{dosovitskiy2021image}
Alexey Dosovitskiy, Lucas Beyer, Alexander Kolesnikov, Dirk Weissenborn, Xiaohua Zhai, Thomas Unterthiner, Mostafa Dehghani, Matthias Minderer, Georg Heigold, Sylvain Gelly, et~al.
\newblock {An image is worth 16x16 words: Transformers for image recognition at scale}.
\newblock {\em arXiv preprint arXiv:2010.11929}, 2020.

\bibitem{dosovitskiy2017carla}
Alexey Dosovitskiy, German Ros, Felipe Codevilla, Antonio Lopez, and Vladlen Koltun.
\newblock {CARLA: An Open Urban Driving Simulator}.
\newblock In {\em Proceedings of the 1st Annual Conference on Robot Learning}. PMLR, 2017.

\bibitem{dutta2022vit}
Pramit Dutta, Ganesh Sistu, Senthil Yogamani, Edgar Galv{\'a}n, et~al.
\newblock {ViT-BEVSeg: A hierarchical transformer network for monocular birds-eye-view segmentation}.
\newblock In {\em 2022 International Joint Conference on Neural Networks (IJCNN)}, pages 1--7. IEEE, 2022.

\bibitem{geiger2012are}
A.~Geiger, P.~Lenz, and R.~Urtasun.
\newblock {Are we ready for autonomous driving? The KITTI vision benchmark suite}.
\newblock In {\em 2012 IEEE Conference on Computer Vision and Pattern Recognition}, page 3354–3361, Providence, RI, Jun 2012. IEEE.

\bibitem{gong2022gitnet}
Shi Gong, Xiaoqing Ye, Xiao Tan, Jingdong Wang, Errui Ding, Yu~Zhou, and Xiang Bai.
\newblock {GitNet: Geometric Prior-based Transformation for Birds-Eye-View Segmentation}.
\newblock {\em arXiv preprint arXiv:2204.07733}, 2022.

\bibitem{gosala2022bird}
Nikhil Gosala and Abhinav Valada.
\newblock {Bird’s-Eye-View Panoptic Segmentation Using Monocular Frontal View Images}.
\newblock {\em IEEE Robotics and Automation Letters}, 7(2), 2022.

\bibitem{harley2022simple}
Adam~W Harley, Zhaoyuan Fang, Jie Li, Rares Ambrus, and Katerina Fragkiadaki.
\newblock {A simple baseline for BEV perception without lidar}.
\newblock {\em arXiv e-prints}, pages arXiv--2206, 2022.

\bibitem{he2015deep}
Kaiming He, Xiangyu Zhang, Shaoqing Ren, and Jian Sun.
\newblock {Deep residual learning for image recognition}.
\newblock In {\em Proceedings of the IEEE conference on computer vision and pattern recognition}, pages 770--778, 2016.

\bibitem{hu2021fiery}
Anthony Hu, Zak Murez, Nikhil Mohan, Sof{\'\i}a Dudas, Jeffrey Hawke, Vijay Badrinarayanan, Roberto Cipolla, and Alex Kendall.
\newblock {FIERY: Future Instance Prediction in Bird's-Eye View From Surround Monocular Cameras}.
\newblock In {\em Proceedings of the IEEE/CVF International Conference on Computer Vision}, pages 15273--15282, 2021.

\bibitem{huang2022bevdet4d}
Junjie Huang and Guan Huang.
\newblock {Bevdet4d: Exploit temporal cues in multi-camera 3d object detection}.
\newblock {\em arXiv preprint arXiv:2203.17054}, 2022.

\bibitem{huang2022bevdet}
Junjie Huang, Guan Huang, Zheng Zhu, and Dalong Du.
\newblock {Bevdet: High-performance multi-camera 3d object detection in bird-eye-view}.
\newblock {\em arXiv preprint arXiv:2112.11790}, 2021.

\bibitem{huang2015densebox}
Lichao Huang, Yi~Yang, Yafeng Deng, and Yinan Yu.
\newblock {Densebox: Unifying landmark localization with end to end object detection}.
\newblock {\em arXiv preprint arXiv:1509.04874}, 2015.

\bibitem{jakobi1995noise}
Nick Jakobi, Phil Husbands, and Inman Harvey.
\newblock {Noise and the reality gap: The use of simulation in evolutionary robotics}.
\newblock In {\em European Conference on Artificial Life}, pages 704--720. Springer, 1995.

\bibitem{klingner2022detecting}
Marvin Klingner, Varun~Ravi Kumar, Senthil Yogamani, Andreas B{\"a}r, et~al.
\newblock {Detecting adversarial perturbations in multi-task perception}.
\newblock In {\em 2022 IEEE/RSJ International Conference on Intelligent Robots and Systems (IROS)}, pages 13050--13057. IEEE, 2022.

\bibitem{kumar2021svdistnet}
Varun~Ravi Kumar, Marvin Klingner, Senthil Yogamani, Markus Bach, et~al.
\newblock {SVDistNet: Self-supervised near-field distance estimation on surround view fisheye cameras}.
\newblock {\em IEEE Transactions on Intelligent Transportation Systems}, 23(8):10252--10261, 2021.

\bibitem{kumar2018near}
Varun~Ravi Kumar, Stefan Milz, Christian Witt, Martin Simon, et~al.
\newblock {Near-field depth estimation using monocular fisheye camera: A semi-supervised learning approach using sparse LiDAR data}.
\newblock In {\em CVPR Workshop}, volume~7, page~2, 2018.

\bibitem{kumar2020unrectdepthnet}
Varun~Ravi Kumar, Senthil Yogamani, Markus Bach, Christian Witt, Stefan Milz, and Patrick M{\"a}der.
\newblock Unrectdepthnet: Self-supervised monocular depth estimation using a generic framework for handling common camera distortion models.
\newblock In {\em 2020 IEEE/RSJ International Conference on Intelligent Robots and Systems (IROS)}, pages 8177--8183. IEEE, 2020.

\bibitem{lang2019pointpillars}
Alex~H Lang, Sourabh Vora, Holger Caesar, Lubing Zhou, Jiong Yang, and Oscar Beijbom.
\newblock {Pointpillars: Fast encoders for object detection from point clouds}.
\newblock In {\em Proceedings of the IEEE/CVF conference on computer vision and pattern recognition}, pages 12697--12705, 2019.

\bibitem{leang2020dynamic}
Isabelle Leang, Ganesh Sistu, Fabian B{\"u}rger, Andrei Bursuc, et~al.
\newblock {Dynamic task weighting methods for multi-task networks in autonomous driving systems}.
\newblock In {\em 2020 IEEE 23rd International Conference on Intelligent Transportation Systems (ITSC)}, pages 1--8. IEEE, 2020.

\bibitem{li2022bevdepth}
Yinhao Li, Zheng Ge, Guanyi Yu, Jinrong Yang, Zengran Wang, Yukang Shi, Jianjian Sun, and Zeming Li.
\newblock {Bevdepth: Acquisition of reliable depth for multi-view 3d object detection}.
\newblock {\em arXiv preprint arXiv:2206.10092}, 2022.

\bibitem{li2022bevformer}
Zhiqi Li, Wenhai Wang, Hongyang Li, Enze Xie, Chonghao Sima, Tong Lu, Qiao Yu, and Jifeng Dai.
\newblock {BEVFormer: Learning Bird's-Eye-View Representation from Multi-Camera Images via Spatiotemporal Transformers}.
\newblock {\em arXiv preprint arXiv:2203.17270}, 2022.

\bibitem{liu2021swin}
Ze~Liu, Yutong Lin, Yue Cao, Han Hu, Yixuan Wei, Zheng Zhang, Stephen Lin, and Baining Guo.
\newblock {Swin transformer: Hierarchical vision transformer using shifted windows}.
\newblock In {\em Proceedings of the IEEE/CVF International Conference on Computer Vision}, pages 10012--10022, 2021.

\bibitem{liu2022bevfusion}
Zhijian Liu, Haotian Tang, Alexander Amini, Xinyu Yang, Huizi Mao, Daniela Rus, and Song Han.
\newblock {BEVFusion: Multi-Task Multi-Sensor Fusion with Unified Bird's-Eye View Representation}.
\newblock {\em arXiv preprint arXiv:2205.13542}, 2022.

\bibitem{lu2019monocular}
Chenyang Lu, Marinus Jacobus~Gerardus van~de Molengraft, and Gijs Dubbelman.
\newblock {Monocular Semantic Occupancy Grid Mapping with Convolutional Variational Encoder-Decoder Networks}.
\newblock {\em IEEE Robotics and Automation Letters}, 4(2), 2019.

\bibitem{ma2022vision}
Yuexin Ma, Tai Wang, Xuyang Bai, Huitong Yang, Yuenan Hou, Yaming Wang, Yu~Qiao, Ruigang Yang, Dinesh Manocha, and Xinge Zhu.
\newblock Vision-centric bev perception: A survey.
\newblock {\em arXiv preprint arXiv:2208.02797}, 2022.

\bibitem{ming2021deep}
Yue Ming, Xuyang Meng, Chunxiao Fan, and Hui Yu.
\newblock Deep learning for monocular depth estimation: A review.
\newblock {\em Neurocomputing}, 438:14--33, 2021.

\bibitem{mohan2021efficientps}
Rohit Mohan and Abhinav Valada.
\newblock {Efficientps: Efficient panoptic segmentation}.
\newblock {\em International Journal of Computer Vision}, 129(5):1551--1579, 2021.

\bibitem{mohapatra2021limoseg}
Sambit Mohapatra, Mona Hodaei, Senthil Yogamani, Stefan Milz, et~al.
\newblock {LiMoSeg: Real-time Bird's Eye View based LiDAR Motion Segmentation}.
\newblock In {\em Proceedings of the 17th International Joint Conference on Computer Vision, Imaging and Computer Graphics Theory and Applications (VISIGRAPP 2022)}, 2022.

\bibitem{pan2019cross}
Bowen Pan, Jiankai Sun, Ho~Yin~Tiga Leung, Alex Andonian, and Bolei Zhou.
\newblock {Cross-view Semantic Segmentation for Sensing Surroundings}.
\newblock {\em IEEE Robotics and Automation Letters}, 5(3), 2019.

\bibitem{philion2020lift}
Jonah Philion and Sanja Fidler.
\newblock {Lift, splat, shoot: Encoding images from arbitrary camera rigs by implicitly unprojecting to 3d}.
\newblock In {\em European Conference on Computer Vision}, pages 194--210. Springer, 2020.

\bibitem{premebida2016high}
Cristiano Premebida, Luis Garrote, Alireza Asvadi, A~Pedro Ribeiro, and Urbano Nunes.
\newblock High-resolution lidar-based depth mapping using bilateral filter.
\newblock In {\em 2016 IEEE 19th international conference on intelligent transportation systems (ITSC)}, pages 2469--2474. IEEE, 2016.

\bibitem{rashed2019motion}
Hazem Rashed, Ahmad El~Sallab, Senthil Yogamani, and Mohamed ElHelw.
\newblock {Motion and depth augmented semantic segmentation for autonomous navigation}.
\newblock In {\em Proceedings of the IEEE/CVF Conference on Computer Vision and Pattern Recognition Workshops}, pages 0--0, 2019.

\bibitem{rashed2021bev}
Hazem Rashed, Mariam Essam, Maha Mohamed, Ahmad~Ei Sallab, and Senthil Yogamani.
\newblock Bev-modnet: Monocular camera based bird's eye view moving object detection for autonomous driving.
\newblock In {\em 2021 IEEE International Intelligent Transportation Systems Conference (ITSC)}, pages 1503--1508. IEEE, 2021.

\bibitem{reading2021categorical}
Cody Reading, Ali Harakeh, Julia Chae, and Steven~L Waslander.
\newblock {Categorical depth distribution network for monocular 3d object detection}.
\newblock In {\em Proceedings of the IEEE/CVF Conference on Computer Vision and Pattern Recognition}, pages 8555--8564, 2021.

\bibitem{reiher2020sim2real}
Lennart Reiher, Bastian Lampe, and Lutz Eckstein.
\newblock {A sim2real deep learning approach for the transformation of images from multiple vehicle-mounted cameras to a semantically segmented image in bird’s eye view}.
\newblock In {\em 2020 IEEE 23rd International Conference on Intelligent Transportation Systems (ITSC)}, pages 1--7. IEEE, 2020.

\bibitem{roddick2020predicting}
Thomas Roddick and Roberto Cipolla.
\newblock {Predicting Semantic Map Representations From Images Using Pyramid Occupancy Networks}.
\newblock In {\em 2020 IEEE/CVF Conference on Computer Vision and Pattern Recognition (CVPR)}, Seattle, WA, USA, 2020. IEEE.

\bibitem{roddick2018orthographic}
Thomas Roddick, Alex Kendall, and Roberto Cipolla.
\newblock {Orthographic feature transform for monocular 3d object detection}.
\newblock {\em arXiv preprint arXiv:1811.08188}, 2018.

\bibitem{saha2021enabling}
Avishkar Saha, Oscar Mendez, Chris Russell, and Richard Bowden.
\newblock {Enabling spatio-temporal aggregation in birds-eye-view vehicle estimation}.
\newblock In {\em 2021 IEEE International Conference on Robotics and Automation (ICRA)}, pages 5133--5139. IEEE, 2021.

\bibitem{saha2022translating}
Avishkar Saha, Oscar Mendez, Chris Russell, and Richard Bowden.
\newblock {Translating images into maps}.
\newblock In {\em 2022 International Conference on Robotics and Automation (ICRA)}, pages 9200--9206. IEEE, 2022.

\bibitem{sistu2019real}
Ganesh Sistu, Isabelle Leang, and Senthil Yogamani.
\newblock {Real-time joint object detection and semantic segmentation network for automated driving}.
\newblock {\em NeurIPSW on ML on the Phone and other Consumer Devices}, 2018.

\bibitem{sun2020scalability}
Pei Sun, Henrik Kretzschmar, Xerxes Dotiwalla, Aurelien Chouard, Vijaysai Patnaik, Paul Tsui, James Guo, Yin Zhou, Yuning Chai, Benjamin Caine, et~al.
\newblock {Scalability in perception for autonomous driving: Waymo open dataset}.
\newblock In {\em Proceedings of the IEEE/CVF conference on computer vision and pattern recognition}, pages 2446--2454, 2020.

\bibitem{tan2020efficientnet}
Mingxing Tan and Quoc Le.
\newblock {Efficientnet: Rethinking model scaling for convolutional neural networks}.
\newblock In {\em International conference on machine learning}, pages 6105--6114. PMLR, 2019.

\bibitem{tan2020efficientdet}
Mingxing Tan, Ruoming Pang, and Quoc~V Le.
\newblock {Efficientdet: Scalable and efficient object detection}.
\newblock In {\em Proceedings of the IEEE/CVF conference on computer vision and pattern recognition}, pages 10781--10790, 2020.

\bibitem{uricar2019challenges}
Michal Uric{\'a}r, David Hurych, Pavel Krizek, and Senthil Yogamani.
\newblock {Challenges in designing datasets and validation for autonomous driving}.
\newblock In {\em Proceedings of the 14th International Joint Conference on Computer Vision, Imaging and Computer Graphics Theory and Applications (VISIGRAPP 2019)}, 2019.

\bibitem{uricar2019desoiling}
Michal Uric{\'a}r, Jan Ulicny, Ganesh Sistu, Hazem Rashed, et~al.
\newblock {Desoiling dataset: Restoring soiled areas on automotive fisheye cameras}.
\newblock In {\em Proceedings of the IEEE/CVF International Conference on Computer Vision Workshops}, pages 0--0, 2019.

\bibitem{vaswani2017attention}
Ashish Vaswani, Noam Shazeer, Niki Parmar, Jakob Uszkoreit, Llion Jones, Aidan~N Gomez, {\L}ukasz Kaiser, and Illia Polosukhin.
\newblock {Attention is all you need}.
\newblock {\em Advances in neural information processing systems}, 30, 2017.

\bibitem{wang2021fcos3d}
Tai Wang, Xinge Zhu, Jiangmiao Pang, and Dahua Lin.
\newblock {Fcos3d: Fully convolutional one-stage monocular 3d object detection}.
\newblock In {\em Proceedings of the IEEE/CVF International Conference on Computer Vision}, pages 913--922, 2021.

\bibitem{wang2021detr3d}
Yue Wang, Vitor~Campagnolo Guizilini, Tianyuan Zhang, Yilun Wang, Hang Zhao, and Justin Solomon.
\newblock {Detr3d: 3d object detection from multi-view images via 3d-to-2d queries}.
\newblock In {\em Conference on Robot Learning}, pages 180--191. PMLR, 2022.

\bibitem{xie2022m2bev}
Enze Xie, Zhiding Yu, Daquan Zhou, Jonah Philion, Anima Anandkumar, Sanja Fidler, Ping Luo, and Jose~M Alvarez.
\newblock {M\^ 2BEV: Multi-Camera Joint 3D Detection and Segmentation with Unified Birds-Eye View Representation}.
\newblock {\em arXiv preprint arXiv:2204.05088}, 2022.

\bibitem{yin2021center}
Tianwei Yin, Xingyi Zhou, and Philipp Krahenbuhl.
\newblock {Center-based 3d object detection and tracking}.
\newblock In {\em Proceedings of the IEEE/CVF conference on computer vision and pattern recognition}, pages 11784--11793, 2021.

\bibitem{yu2018deep}
Fisher Yu, Dequan Wang, Evan Shelhamer, and Trevor Darrell.
\newblock {Deep layer aggregation}.
\newblock In {\em Proceedings of the IEEE conference on computer vision and pattern recognition}, pages 2403--2412, 2018.

\bibitem{zhao2017pyramid}
Hengshuang Zhao, Jianping Shi, Xiaojuan Qi, Xiaogang Wang, and Jiaya Jia.
\newblock {Pyramid scene parsing network}.
\newblock In {\em Proceedings of the IEEE conference on computer vision and pattern recognition}, pages 2881--2890, 2017.

\bibitem{zhou2022cross}
Brady Zhou and Philipp Kr{\"a}henb{\"u}hl.
\newblock {Cross-view Transformers for real-time Map-view Semantic Segmentation}.
\newblock In {\em Proceedings of the IEEE/CVF Conference on Computer Vision and Pattern Recognition}, pages 13760--13769, 2022.

\end{thebibliography}

\printindex
\cleardoublepage
\end{document}